\documentclass[letterpaper]{article}
\usepackage{aaai20}
\usepackage{times}
\usepackage{helvet}
\usepackage{courier}
\usepackage[hyphens]{url}
\usepackage{graphicx}
\usepackage{graphicx}
\usepackage{amsmath}
\usepackage{amssymb}

\usepackage{booktabs}

\usepackage{algorithm}
\usepackage{algpseudocode}

\title{Cakewalk Sampling}
\author {Uri Patish, Shimon Ullman\\
	Weizmann Institute of Science, Rehovot, Israel}

\begin{document}
\maketitle
\begin{abstract}
We study the task of finding good local optima in combinatorial optimization
problems. Although combinatorial optimization is NP-hard in general,
locally optimal solutions are frequently used in practice. Local search
methods however typically converge to a limited set of optima that
depend on their initialization. Sampling methods on the other hand
can access any valid solution, and thus can be used either directly
or alongside methods of the former type as a way for finding good
local optima. Since the effectiveness of this strategy depends on
the sampling distribution, we derive a robust learning algorithm that
adapts sampling distributions towards good local optima of arbitrary
objective functions. As a first use case, we empirically study the
efficiency in which sampling methods can recover locally maximal cliques
in undirected graphs. Not only do we show how our adaptive sampler
outperforms related methods, we also show how it can even approach
the performance of established clique algorithms. As a second use
case, we consider how greedy algorithms can be combined with our adaptive
sampler, and we demonstrate how this leads to superior performance
in k-medoid clustering. Together, these findings suggest that our
adaptive sampler can provide an effective strategy to combinatorial
optimization problems that arise in practice.
\end{abstract}

\section{\label{sec:Introduction}Introduction}

Combinatorial optimization is one of the foundational problems of
computer science. Though in general such problems are NP-hard \cite{Papadimitriou2003},
it is often the case that locally optimal solutions can be useful
in practice. In clustering for example, a common objective is to divide
a given set of examples into a fixed number of groups so as to minimize
the distances between group members. Since enumerating all the possible
groupings is usually intractable, local search methods such as k-means
\cite{Macqueen1967} are frequently used to approach such problems.
In many problems however, methods that transform one solution to another
can be highly sensitive to their initialization. In some cases this
is a result of applying a local search to a problem which has multiple
local optima. In others, the search space is simply disconnected,
and transforming one valid solution to another is only possible within
a small sub-space. In such cases, the quality of the final solution
is determined by how the search is initialized. One common heuristic
around this is to sample few initial solutions, and to apply the search
multiple times. However, the success of this heuristic mostly depends
on the sampling distribution that produces these initial solutions.
Thus, if we can have an algorithm that adapts a sampling distribution
towards solutions that are associated with good objective values,
then we might be able to use it to find good local optima. Such a
method could potentially sample locally optimal solutions on its own,
or be used as an algorithm that learns how to initialize a particular
local search.

One type of algorithms which seem suitable for the task, and which
have drawn considerable interest in the last few years are policy
gradient methods \cite{Sutton2017}. Such methods construct a parametric
sampling distribution over the search space, and optimize the expected
value of some objective function by applying gradient updates in the
parameters' space. On the surface, when provided with the right sampling
distribution, such methods can access any valid solution, and can
therefore provide a strategy that is suitable to our setting. Nonetheless,
a closer inspection reveals these methods are highly sensitive to
perturbations of the objective function. In particular, the objective
values directly affect the sign and the magnitude of the gradient,
making these methods notoriously hard to tune. Since the objective
in this construction is essentially a random variable whose distribution
changes from problem to problem, finding a general rule for tuning
them seems impractical. Following this understanding, we propose to
circumvent this sensitivity by utilizing a generic surrogate objective
function that has the following two properties. First, the surrogate
should preserve the set of locally optimal solutions. Second, the
surrogate should have a predetermined distribution for every possible
objective. Once in this form, such constructions can provide us with
a generic adaptive sampler. With this idea in mind, we show how the
empirical cumulative distribution function (CDF henceforth) of the
original objective can be used to construct such surrogate objectives,
and we present a version which makes the basis of our method. Since
the crux of our method is based on capitalizing on the CDF of the
original objective, we refer to our method as CAkEWaLK which stands
for CumulAtivEly Weighted LiKelihood.

We start by considering adaptive sampling methods for combinatorial
optimization problems in section \ref{sec:Background}, and proceed
to present Cakewalk in section \ref{sec:Cakewalk}. In section \ref{subsec:Optimization-Related-Work}
we discuss how Cakewalk is related to policy-gradient methods in reinforcement
learning, to multi-arm bandit algorithms, and to the cross-entropy
(CE henceforth) method. Since ideally we would like to have an adaptive
sampler which can recover locally optimal solutions on its own, we
use the problem of finding inclusion maximal cliques in undirected
graphs as a controlled experiment for testing this property in a non-trivial
setting. For that matter, in section \ref{sec:Application-to-Maximum-Clique}
we investigate how to apply such methods to the clique problem, and
we report experimental results on a dataset of 80 graphs that is regularly
used as a benchmark for clique algorithms. In section \ref{sec:Application-to-K-Medoids}
we consider how Cakewalk can be combined with greedy algorithms, and
we demonstrate such a use case on k-medoid clustering, the combinatorial
counterpart of k-means. We test how Cakewalk compares to two greedy
algorithms commonly used to approach the problem on 38 small datasets,
and we show how using Cakewalk for learning how to initialize these
methods produced the best performance. We then conclude with a few
final remarks in section \ref{sec:Conclusion}.

\section{\label{sec:Background}Background}

We construct an adaptive sampler for combinatorial optimization problems,
and start by stating the problem. Let $f$ be an objective function
which we need to maximize, and let $\boldsymbol{x}\in\left[M\right]{}^{N}$
be a string that describes $N$ items such that each $x_{j}$ is one
of a discrete set of $M$ items. In this text we denote discrete sets
$\left\{ 1,\dots,K\right\} $ using $\left[K\right]$. Our goal is
to search a possibly constrained space $\mathcal{X}\subseteq\left[M\right]{}^{N}$
for some $\boldsymbol{x}^{*}$ that achieves an optimal $f\left(\boldsymbol{x}^{*}\right)=y^{*}$
(for constrained problems $\mathcal{X}\subsetneq\left[M\right]{}^{N}$).
Since $\mathcal{X}$ is discrete and high-dimensional, in general
this problem is NP-hard (maximum clique can be reduced to this description),
hence we focus only on finding locally optimal solutions. For the
purpose of defining locally optimal solutions, we rely on a neighborhood
function $\mathcal{N}$ that maps each $\boldsymbol{x}$ to its neighboring
set. For example, if $\mathcal{X}=\left\{ 0,1\right\} ^{N}$, then
a neighborhood function could be $\mathcal{N}\left(\boldsymbol{x}\right)=\left\{ \boldsymbol{x}^{\prime}\in\mathcal{X}\vert\sum_{i=1}^{N}\left|x_{i}-x_{i}^{\prime}\right|=1\right\} $.
Note however that the methods we describe treat $f$ as a black-box,
and do not require $\mathcal{N}$ for their operation. Our goal is
to find some locally optimal solution $\boldsymbol{x}^{*}\in\mathcal{X}_{f}^{*}$
where the set of locally optimal solutions is defined as $\mathcal{X}_{f}^{*}=\left\{ \boldsymbol{x}\in\mathcal{X}\vert\forall\boldsymbol{x}'\in\mathcal{N}\left(\boldsymbol{x}\right).f\left(\boldsymbol{x}\right)\ge f\left(\boldsymbol{x}'\right)\right\} $.
Preferably, we would like to find some $\boldsymbol{x}^{*}$ whose
objective value $y^{*}=f\left(\boldsymbol{x}^{*}\right)$ is as large
as possible, though in general, this cannot be guaranteed.

We describe a learning algorithm for problems of this type. Let $\boldsymbol{X}$
be a random variable that is defined over $\mathcal{X}$, and which
is distributed according to a parametric distribution $\mathbb{P}_{\boldsymbol{\theta}}$
that the algorithm maintains. In addition, let $\boldsymbol{Y}$ be
a random variable that is defined over the values of the objective
function $f$, i.e. $\boldsymbol{Y}=f\left(\boldsymbol{X}\right)$.
We emphasize that in this text we refer to random variables using
capital English letters in bold such as $\boldsymbol{X}$ or $\boldsymbol{Y}$,
and we use $\boldsymbol{x}$ and $y$ to refer to elements in their
appropriate sample spaces (deterministic quantities). The algorithm
we describe iteratively samples solutions according to $\mathbb{P}_{\boldsymbol{\theta}}$,
and it updates the parameters $\boldsymbol{\theta}\in\mathbb{R}^{d}$
which govern $\mathbb{P}_{\boldsymbol{\theta}}$ in a manner that
reflects the quality of those solutions. Initially $\mathbb{P}_{\boldsymbol{\theta}}$
is set to have high entropy, but as the algorithm progresses, the
entropy in the distribution is decreased until eventually only few
solutions become likely (for a discussion of entropy as measure of
uncertainty see \cite{Cover2012}). At this point, sampling some $\boldsymbol{x}$
from $\mathbb{P}_{\boldsymbol{\theta}}$ should return some locally
optimal solution with high probability. Since we discuss an iterative
algorithm that at each iteration $t$ updates the parameters $\boldsymbol{\theta}_{t}$,
we refer to the random variables that are associated with $\mathbb{P}_{\boldsymbol{\theta}_{t}}$
by $\boldsymbol{X}_{t}$ and $\boldsymbol{Y}_{t}$. Lastly, as a short
hand notation, we refer to $\mathbb{P}_{\boldsymbol{\theta}}\left(\boldsymbol{X}=\boldsymbol{x}\right)$
simply by $\mathbb{P}_{\boldsymbol{\theta}}\left(\boldsymbol{x}\right)$.

Since we learn a distribution function, we say that our learning objective
$J$$\left(\boldsymbol{\theta}\right)$ is to maximize the expectation
over $\boldsymbol{x}\sim\mathbb{P}_{\boldsymbol{\theta}}$ of the
original objective which we denote as $\mathbb{E}_{\boldsymbol{\theta}}\left[\boldsymbol{Y}\right]$.
To find the parameters $\boldsymbol{\theta}$ which maximize $J\left(\boldsymbol{\theta}\right)=\mathbb{E}_{\boldsymbol{\theta}}\left[\boldsymbol{Y}\right]$,
we derive a gradient ascent algorithm which relies on estimates of
$\nabla_{\boldsymbol{\theta}}\mathbb{E}_{\boldsymbol{\theta}}\left[\boldsymbol{Y}\right]$.
To calculate the gradient, we use the log-derivative trick, $\nabla_{\boldsymbol{\theta}}\mathbb{E}_{\boldsymbol{\theta}}\left[\boldsymbol{Y}\right]=\mathbb{E}_{\boldsymbol{\theta}}\left[\boldsymbol{Y}\nabla_{\boldsymbol{\theta}}\log\mathbb{P}_{\boldsymbol{\theta}}\left(\boldsymbol{X}\right)\right]$,
which allows us to estimate $\mathbb{E}_{\boldsymbol{\theta}}\left[\boldsymbol{Y}\nabla_{\boldsymbol{\theta}}\log\mathbb{P}_{\boldsymbol{\theta}}\left(\boldsymbol{X}\right)\right]$
through Monte Carlo sampling \cite{Wasserman2013}. Traditionally,
at each iteration $t$, a large sample $S_{t}=\left\{ \boldsymbol{x}_{t}^{k},y_{t}^{k}\right\} _{k=1}^{K}$
of some fixed size $K$ is sampled using $\mathbb{P}_{\boldsymbol{\theta}_{t}}$.
Denoting this estimate by $\boldsymbol{\Delta}_{t}$, then the update
at iteration $t$ takes the following form, 
\begin{align}
\boldsymbol{\theta}_{t} & =\boldsymbol{\theta}_{t-1}+\eta_{t}\boldsymbol{\Delta}_{t}\label{eq:parameters-update}\\
\boldsymbol{\Delta}_{t} & =\frac{1}{K}\sum_{k=1}^{K}\left[y_{t}^{k}\nabla_{\boldsymbol{\theta}}\log\mathbb{P}_{\boldsymbol{\theta}}\left(\boldsymbol{x}_{t}^{k}\right)\right]\label{eq:estimated-gradient}
\end{align}
\\
where $\eta_{t}$ is a positive learning rate parameter that is predetermined.
We describe the update step using a vanilla gradient update mostly
for illustratory purposes, though in practice any gradient based update
such as Adam \cite{Kingma2014} or AdaGrad \cite{Duchi2011} can be
used instead. 

While this stochastic optimization scheme can theoretically converge
to a local maximum of $J$ \cite{Williams1992}, in practice it is
highly sensitive to choices of $K$ and $\left\{ \eta_{t}\right\} _{t=1}^{T}$,
and to the distributions of $\left\{ \boldsymbol{Y}_{t}\right\} _{t=1}^{T}$
(exemplified in the next section). One way to handle this sensitivity
is to draw large samples in each iteration, which can reduce the variance
of the gradient estimator (in the combinatorial setting, this might
require exponentially sized samples). However, even if we increase
the sample size, we still need to find a rule that adjusts $\eta_{t}$
to the distribution of $Y_{t}$ if we are to produce a generic sampler.
Thus, we approach this problem differently, and consider instead how
can we adjust the distribution of the objective regardless of the
sample size. We focus on online updates (setting $K=1$), and accordingly
drop the superscript $k$ when referring to $\boldsymbol{x}_{t}^{k}$
and $y_{t}^{k}$ for the remainder of the text.

\section{\label{sec:Cakewalk}Cakewalk}

We start by examining equations \ref{eq:parameters-update} and \ref{eq:estimated-gradient},
and observing that we update $\boldsymbol{\theta}_{t}$ by making
a step $\eta_{t}y_{t}$ in the gradient's direction $\nabla_{\boldsymbol{\theta}}\log\mathbb{P}_{\boldsymbol{\theta}}\left(\boldsymbol{x}_{t}\right)$.
Thus, the sign and magnitude of $\eta_{t}y_{t}$ essentially determine
whether we increase or decrease the (log) likelihood of $\boldsymbol{x}_{t}$,
and to what extent we do so. Such direct dependence on the objective
values could make our sampler susceptible to perturbations of the
objective function. For example, suppose that we have two functions
such that $f_{2}\left(\boldsymbol{x}\right)=cf_{1}\left(\boldsymbol{x}\right)$
for every $\boldsymbol{x}$, with $c$ being some fixed positive constant.
Clearly, $\mathcal{X}_{f_{1}}^{*}=\mathcal{X}_{f_{2}}^{*}$, nonetheless,
sampling and updating the parameters using equations \ref{eq:parameters-update}
and \ref{eq:estimated-gradient} would change the magnitude of the
gradient updates by a factor $c$. Though one can adjust the learning
rates to the particularities of some given objective, such an approach
would require that we tune our method on a case by case basis. 

More generally, it appears that the distributions of $\left\{ \eta_{t}\boldsymbol{Y}_{t}\right\} _{t=1}^{T}$
play a critical role in our gradient process. If for example $\left|\eta_{t}\boldsymbol{Y}_{t}\right|$
is unbounded from above for all $t$, we might take steps that are
too large which may cause the gradient process to diverge. Steps that
are too small are unfavorable as well, as these will maintain too
much entropy in $\mathbb{P}_{\theta},$ and due to the discrete nature
of $\mathcal{X}$, finding good $\boldsymbol{x}$s can take exponentially
many examples. Since in general we do not know ahead of time the distribution
of each $\boldsymbol{Y}_{t}$, if we follow the construction presented
in section \ref{sec:Background}, we will not be able to determine
the series $\left\{ \eta_{t}\right\} _{t=1}^{T}$ in a manner that
would fit all scenarios. This reasoning leads us to conclude that
if we wish to obtain generic updates, we must come up with some surrogate
objective function which preserves $\mathcal{X}_{f}^{*}$, and for
which we can determine the distributions of $\left\{ \boldsymbol{Y}_{t}\right\} _{t=1}^{T}$
ahead of time. To that end, we introduce a weight function $w$ that
when composed over $f$ (i.e. $w\circ f$) produces a surrogate objective
that meets these criteria. 

Preserving the original set of optimal solutions is the easy part,
as we just need to require that $w$ will be monotonically increasing,
and that would imply that $\mathcal{X}_{f}^{*}\subseteq\mathcal{X}_{w\circ f}^{*}$
(and strict monotonicity would ensure that $\mathcal{X}_{f}^{*}=\mathcal{X}_{w}^{*}$).
The harder part is to construct $w$ in a manner that would fix the
distribution of $w$$\left(\boldsymbol{Y}_{t}\right)$ for all $t$.
Nonetheless, basic probability tells us that if $F_{t}$ is the CDF
of $Y_{t}$, then $F_{t}\left(\boldsymbol{Y}_{t}\right)$ is uniformly
distributed on $\left[0,1\right]$ \cite{Wasserman2013}. Since every
CDF is monotonic increasing, if we construct $w$ using $F_{t}$,
we can preserve the original set of optimal solutions. More importantly,
if we can estimate $F_{t}$, we could use it to produce our surrogate
objective as it \textbf{would fix the surrogate's distribution once
and for all}, thus making significant progress towards our goal. Since
$w\left(Y_{t}\right)\sim U\left(0,1\right)$ might not be ideal, we
can utilize another monotonic increasing function $g$ for which $g\left(F_{t}\left(\boldsymbol{Y}_{t}\right)\right)$
can be distributed differently. For purposes that we specify next,
we also require that $g$ will be bounded.

Since we do not have access to $F_{t}$ in general, as was the case
with the gradient, we need to estimate it from data. Fortunately enough,
since the image of $f$ is one dimensional (an optimization objective),
order statistics can supply us with highly reliable non-parametric
estimates for each $F_{t}$. At this point however, it is worth considering
how can we estimate $F_{t}$ without drawing a large sample at each
iteration. Due to equation \ref{eq:estimated-gradient}, if we use
a sampling distribution for which $\left\Vert \nabla_{\boldsymbol{\theta}}\log\mathbb{P}_{\boldsymbol{\theta}}\left(\boldsymbol{x}_{t}\right)\right\Vert $
is bounded, then since $w\left(y_{t}\right)$ is bounded as well,
$\left\Vert \boldsymbol{\Delta}_{t}\right\Vert $ will be bounded
for every $\boldsymbol{x}_{t}$ and $y_{t}$. This implies that we
can control how different the parameters will be between any two iterations:
for any two iterations $t$ and $t-k$ where $k\in\left[t-1\right]$,
we can make $\left\Vert \boldsymbol{\theta}_{t}-\boldsymbol{\theta}_{t-k}\right\Vert $
as small as we want simply by changing $\eta_{t}$. Thus, instead
of drawing a large sample in each iteration, we can say the last objective
values $y_{t-1},\dots,y_{t-k}$ are approximately i.i.d from $\mathbb{P}_{\boldsymbol{\theta}_{t-1}}$.
Therefore, if we use small enough learning rates, we can use $\hat{F}_{t-1}\left(y\right)=\frac{1}{k}\sum_{i=1}^{k}\mathbb{I}\left[y_{t-i}<y\right]$
as an estimator for $F_{t-1}$, where $\mathbb{I}\left[\cdot\right]$
is the indicator function. In our experiments, using some fixed learning
rate $\eta\in\left(0,1\right)$ along with $k=\frac{1}{\eta}$ seem
to work quite well. Overall, the updates we suggest have the following
form,

\begin{equation}
\boldsymbol{\Delta}_{t}=g\left(\hat{F}_{t-1}\left(y_{t}\right)\right)\nabla_{\boldsymbol{\theta}}\log\mathbb{P}_{\boldsymbol{\theta}}\left(\boldsymbol{x}_{t}\right)\label{eq:CAkEWaLK-update}
\end{equation}

\subsection{\label{subsec:Weight-Functions}Surrogate Objectives}

In this section we focus on a single iteration $t$, and thus, drop
the subscript $t$ when discussing two possible weight functions.
One simple option is to use the empirical CDF $\hat{F}$ directly,
which would make $\hat{F}\left(\boldsymbol{Y}\right)$ uniform discrete
on $\left[0,1\right]$. However, this surrogate has a major drawback:
it leads to an increase in the likelihood of every example it sees.
This creates a bias towards $\boldsymbol{x}$s that have already been
sampled, compared with $\boldsymbol{x}$s that were not, even though
their associated objective value might be better. Since $\mathcal{X}$
grows exponentially with $N$, examples that are drawn early in the
process can influence the course of the optimization dramatically.
Following this reasoning, we adjust $\hat{F}$ so that it would increase
the likelihood of only half of the examples, and decrease the likelihood
of the other half. To do so, we make $\hat{w}\left(y\right)=2\hat{F}\left(y\right)-1$.
By construction, it follows that $\hat{w}\left(\boldsymbol{Y}\right)$
is uniform discrete on $\left[-1,1\right]$. In this fashion, when
applied with some fixed learning rate, $\hat{w}$ determines whether
the likelihood of some example will be increased or decreased, and
to what extent. Notably, this is achieved along with full specification
of the distribution of $\hat{w}\left(\boldsymbol{Y}\right)$. This
is a major advantage compared with, for example, transforming $\boldsymbol{Y}$
with its estimated z-score, as in this case we cannot determine how
$w\left(\boldsymbol{Y}\right)$ is distributed, nor can we guarantee
that $\left|w\left(\boldsymbol{Y}\right)\right|$ is bounded (leading
to a risk of divergence, and disrupting the online estimation of $\hat{w}$).
We summarize Cakewalk with $\hat{w}$, and any gradient addition rule $Add$ (this includes hyper-parameters) in algorithm \ref{alg:Cakewalk}.
\begin{algorithm}[tb]
	\caption{Cakewalk}
	\label{alg:Cakewalk}
	\begin{algorithmic}
		\State {\bfseries input} $f$, $\mathbb{P}_{\boldsymbol{\theta}}$, $k$, $Add$
		\Comment {objective function $f$, sampling distribution $\mathbb{P}_{\boldsymbol{\theta}}$, integer $k$, gradient addition rule $Add$}
		\State initialize $\boldsymbol{\theta}_0$
		\While	{not converged, $t = 1,2,\dots$} 
		\State $\boldsymbol{x}_t\sim\mathbb{P}_{\boldsymbol{\theta}_{t-1}}$ \Comment{sampling an example}
		\State $y_t=f\left(\boldsymbol{x}_t\right)$ \Comment{objective value}
		\If {$t>k$}
		\State $w_{t}=2\left(\frac{1}{k}\sum_{i=1}^{k}\mathbb{I}\left[y_{t-i}<y_{t}\right]\right)-1$
		\State $\boldsymbol{\Delta}_{t} =w_{t}\nabla_{\boldsymbol{\theta}}\log\mathbb{P}_{\boldsymbol{\theta}}\left(\boldsymbol{x}_{t}\right)$			
		\State $\boldsymbol{\theta}_{t} = Add\left(\boldsymbol{\theta}_{t-1},\boldsymbol{\Delta}_{t}\right)$
		\EndIf
		\EndWhile
		\State {\bfseries return} $\boldsymbol{x^{*}}$ which had the highest $y^{*}$
	\end{algorithmic}
\end{algorithm}

\section{\label{subsec:Optimization-Related-Work}Related Work}

Cakewalk is closely related to policy gradient methods. The research
on these methods was initiated by Williams with REINFORCE \cite{Williams1988},
an algorithm which we consider as the prototype to Cakewalk, and which
provides Cakewalk with convergence guarantees. Most of the work on
policy gradient methods derives from REINFORCE, essentially discussing
how to transform the objective in various scenarios. Most commonly
these involve a baseline estimate $\hat{\mu}$ of $\mathbb{E}\left(\boldsymbol{Y}\right)$
that can be used to make $\mathbb{E}\left(\boldsymbol{Y}-\hat{\mu}\right)=0$,
or a problem specific model for $\boldsymbol{Y}$ as is done in actor-critic
methods \cite{Sutton2017}. While sometimes useful, in these constructions
the distribution of the objective usually remains unknown, and as
a result these methods require careful tuning that is mostly done
through trial and error. Cakewalk on the other hand uses a surrogate
objective whose distribution is predetermined, and therefore can be
applied to a variety of problems in the same way. 

Cakewalk is also reminiscent of multi-arm bandit algorithms. In the
bandit setting, a learner is faced with a sequential decision problem,
where in each round an arm is chosen, and each arm is associated with
some non-deterministic loss. Initially suggested by Thompson \cite{Thompson1933},
this setting has been explored extensively with the notable successes
of the UCB algorithm \cite{Auer2002UCB-a,Auer2002UCB-b} for cases
where the losses are stochastic, and Exp3 \cite{Auer1995,Auer2002Exp3}
for when they can even be determined by an adversary. Over the years
these have become a basis for a wide variety of algorithms \cite{Bubeck2012}
for various settings which even extend to cases that involve high
dimensional structured arms \cite{Awerbuch2004,Mcmahan2004,CesaBianchi2012}.
The key difference between the bandit and the optimization setting
is that the losses associated with each of the arms are non-deterministic,
and thus in the bandit setting the main challenge is to balance estimating
the statistics associated with each of the arms, with exploiting the
information that was already gathered. In the optimization setting
however, the goal is to find the best deterministic solution using
the least number of steps. Thus, in spite of the apparent similarity,
it is this fundamental difference that separates the optimization
from bandit settings, and which leads to fundamentally different algorithms. 

Cakewalk is also related to the CE method, an iterative algorithm
for adapting an importance sampler towards some event of interest.
CE was initially introduced by Rubinstein for estimating low probability
events \cite{Rubinstein1997}, and later adapted to combinatorial
optimization problems \cite{Rubinstein2001}. In CE, at each iteration
a large set of examples are sampled. Then, the examples are sorted
according to some performance measure (objective function in optimization),
and the examples that belong to some highest percentile are selected.
For the update step, the distribution's parameters are set to the
maximum likelihood estimate of the selected examples. In this sense,
CE is similar to Cakewalk as both methods are only sensitive to the
objective values order. However, CE requires large samples at each
iteration, and it requires distributions for which maximum likelihood
estimates can be produced efficiently (usually this means in closed
form). Cakewalk on the other hand only requires a single example at
each iteration, and can be applied with any differentiable sampling
distribution. Thus, not only can Cakewalk be considerably less costly
than CE, its potential applications are much broader.

\section{\label{sec:Application-to-Maximum-Clique}Maximum Clique}

\begin{table*}[t]
	\caption{Rate of locally optimal solutions, higher is better}
	\label{tbl:Local-Optimality-SCS}
	\centering
	\begin{tabular}{lllllllll}
		\toprule
		 & $\text{Exp3}$ & $\text{REINF}$ & $\text{REINF}_{B}$ & $\text{REINF}_{Z}$ & $\text{OCE}_{0.01}$ & $\text{OCE}_{0.1}$ & $\text{CW}\,\hat{F}$ & $\text{CW}\,\hat{w}$ \\
		\midrule
		SGA & $\text{0.000}^{*}$ & $\text{0.001}^{*}$ & $\text{0.001}^{*}$ & $\text{0.097}$ & $\text{0.001}^{*}$ & $\text{0.000}^{*}$ & $\text{0.002}^{*}$ & $\text{0.042}$\\
		AdaGrad & $\text{0.000}^{*}$ & $\text{0.000}^{*}$ & $\text{0.352}^{*}$ & $\text{0.427}^{*}$ & $\text{0.077}^{*}$ & $\text{0.691}^{*}$ & $\text{0.164}^{*}$ & \textbf{0.835}\\
		Adam & $\text{0.000}^{*}$ & $\text{0.000}^{*}$ & $\text{0.525}^{*}$ & $\text{0.616}^{*}$ & $\text{0.106}^{*}$ & $\text{0.353}^{*}$ & $\text{0.184}^{*}$ & $\text{0.753}$\\
		\bottomrule
	\end{tabular}
\end{table*}

\begin{table*}[t]
	\caption{Rate of inclusion maximal cliques, higher is better}
	\label{tbl:Local-Optimality-IncMax}
	\centering
	\begin{tabular}{lllllllll}
		\toprule
		 & $\text{Exp3}$ & $\text{REINF}$ & $\text{REINF}_{B}$ & $\text{REINF}_{Z}$ & $\text{OCE}_{0.01}$ & $\text{OCE}_{0.1}$ & $\text{CW}\,\hat{F}$ & $\text{CW}\,\hat{w}$ \\		\midrule
		SGA & $\text{0.000}$ & $\text{0.000}$ & $\text{0.000}$ & $\text{0.138}$ & $\text{0.000}$ & $\text{0.000}$ & $\text{0.000}$ & $\text{0.037}$\\
		AdaGrad & $\text{0.000}^{*}$ & $\text{0.000}^{*}$ & $\text{0.637}^{*}$ & $\text{0.688}^{*}$ & $\text{0.063}^{*}$ & $\text{0.875}$ & $\text{0.175}^{*}$ & \textbf{0.912}\\
		Adam & $\text{0.000}^{*}$ & $\text{0.000}^{*}$ & $\text{0.662}^{*}$ & $\text{0.787}$ & $\text{0.100}^{*}$ & $\text{0.412}^{*}$ & $\text{0.212}^{*}$ & $\text{0.887}$\\
		\bottomrule
	\end{tabular}
\end{table*}

\begin{table*}[t]
	\caption{Best-sample to total-samples ratio, lower is better}
	\label{tbl:Best-Sample-Ratio}
	\centering
	\begin{tabular}{lllllllll}
		\toprule
		 & $\text{Exp3}$ & $\text{REINF}$ & $\text{REINF}_{B}$ & $\text{REINF}_{Z}$ & $\text{OCE}_{0.01}$ & $\text{OCE}_{0.1}$ & $\text{CW}\,\hat{F}$ & $\text{CW}\,\hat{w}$ \\
		\midrule
		SGA & - & - & $\text{0.654}$ & $\text{0.907}$ & $\text{0.874}$ & $\text{0.945}$ & $\text{0.939}$ & $\text{0.927}$\\
		AdaGrad & - & - & $\text{0.821}^{*}$ & $\text{0.821}^{*}$ & $\text{0.966}^{*}$ & $\text{0.820}^{*}$ & $\text{0.926}^{*}$ & $\text{0.657}$\\
		Adam & - & - & $\text{0.743}^{*}$ & $\text{0.731}^{*}$ & $\text{0.835}^{*}$ & $\text{0.697}^{*}$ & $\text{0.741}^{*}$ & \textbf{0.619}\\
		\bottomrule
	\end{tabular}
\end{table*}

\begin{table*}[t]
	\caption{Largest-returned-clique to largest-known-clique ratio, higher is better}
	\label{tbl:Larget-Clique-Ratio}
	\centering	
	\begin{tabular}{lllllllll}
		\toprule
		 & $\text{Exp3}$ & $\text{REINF}$ & $\text{REINF}_{B}$ & $\text{REINF}_{Z}$ & $\text{OCE}_{0.01}$ & $\text{OCE}_{0.1}$ & $\text{CW}\,\hat{F}$ & $\text{CW}\,\hat{w}$ \\
		\midrule
		SGA & $\text{0.000}$ & $\text{0.000}$ & $\text{0.000}$ & $\text{0.135}$ & $\text{0.000}$ & $\text{0.000}$ & $\text{0.000}$ & $\text{0.038}$\\
		AdaGrad & $\text{0.000}^{*}$ & $\text{0.000}^{*}$ & $\text{0.538}^{*}$ & $\text{0.567}^{*}$ & $\text{0.062}^{*}$ & $\text{0.738}$ & $\text{0.161}^{*}$ & \textbf{0.756}\\
		Adam & $\text{0.000}^{*}$ & $\text{0.000}^{*}$ & $\text{0.577}$ & $\text{0.657}$ & $\text{0.091}^{*}$ & $\text{0.364}^{*}$ & $\text{0.190}^{*}$ & $\text{0.737}$\\
		\bottomrule
	\end{tabular}
\end{table*}

In this section, we study whether adaptive samplers
can recover locally optimal solutions. We emphasize that our goal
in this section is mostly to investigate this question, rather than
compete with iterative algorithms that transform solutions, and search
the input space directly. We study this question on a NP-hard problem
instead of problem in which the global optimum can be found in polynomial
time, as it is important to verify that such methods can recover non-trivial
optima in challenging scenarios. We focus on the problem of finding
inclusion maximal cliques, as the notion of inclusion maximal cliques
naturally entails what neighborhood function should be used to judge
local optimality. Formally, a graph $G$ is a pair $\left(V,E\right)$
where $V=\left[N\right]$ is a set of vertices, and $E\subseteq V\times V$
is a set of edges. $G$ is undirected if for every $\left(i,j\right)\in E$
it follows that $\left(j,i\right)\in E$. A clique in an undirected
graph is a subset of vertices $U\subseteq V$ such that each pair
of which is connected by an edge. An inclusion maximal clique $U$
is such that there is no other $v\in V\setminus U$ for which $U\cup\left\{ v\right\} $
is also a clique. 

We design an objective that could inform algorithms that only rely
on function evaluations how densely connected is some subgraph, and
which favors larger subgraphs. We refer to this objective as the soft-clique-size
function, and denote it by $f_{SCS}$. For our purposes, we say the
space $\mathcal{X}=\left\{ 0,1\right\} ^{N}$ correspond to strings
which determine membership in some subgraph $U$. Let $\boldsymbol{x}\in\mathcal{X}$,
then for each vertex $j\in V$, we say that $j\in U$ if and only
if $x_{j}=1$, and accordingly we denote such subgraphs by $U_{\boldsymbol{x}}$.
If some $U_{\boldsymbol{x}}$ is a clique, for every $i,j\in U_{\boldsymbol{x}},i\neq j$
it follows that $\left(i,j\right)\in E$, and therefore $\underset{i,j\in U_{\boldsymbol{x}},i\neq j}{\sum}\mathbb{I}\left[\left(i,j\right)\in E\right]=\left|U_{\boldsymbol{x}}\right|\left(\left|U_{\boldsymbol{x}}\right|-1\right)$.
As a consequence, for a general subgraph $U_{\boldsymbol{x}}$ dividing
the LHS by the RHS produces a subgraph density term. However, simply
returning a density term would not indicate to an algorithm it should
prefer larger subgraphs over smaller ones. Accordingly, we add a parameter
$\kappa\in\left[0,1\right]$ that rewards larger subgraphs, and we
change the denominator to $\left|U_{\boldsymbol{x}}\right|\left(\left|U_{\boldsymbol{x}}\right|-1+\kappa\right)$.
To see why higher $\kappa$ can reward larger cliques we focus on
the case that $\left|U_{\boldsymbol{x}}\right|\ge2$, and observe
that for $U_{\boldsymbol{x}}$ which is clique our subgraph density
term will be $1$ when $\kappa=0$ . However, when $\kappa=1$, the
subgraph density will be $\frac{\left|U_{\boldsymbol{x}}\right|\left(\left|U_{\boldsymbol{x}}\right|-1\right)}{\left|U_{\boldsymbol{x}}\right|^{2}}=\frac{\left|U_{\boldsymbol{x}}\right|-1}{\left|U_{\boldsymbol{x}}\right|}$,
and thus, the larger $U_{\boldsymbol{x}}$ is, the closer this ratio
is to $1$. In this manner, increasing $\kappa$ gives larger subgraphs
a 'boost' compared to smaller ones. This of course comes at a price,
as it could be that some subgraph which is not a clique will have
a higher score than some smaller subgraph which is a clique (only
for $\kappa=0$ a score of $1$ necessarily means that $U_{\boldsymbol{x}}$
is clique). Empirically, we see that the algorithms we have tested
are not very sensitive to the value of $\kappa$. Lastly, to avoid
division by zero for cases $\left|U_{\boldsymbol{x}}\right|<2$, we
wrap the denominator with $\max\left(\cdot,1\right)$. Altogether,
our soft-clique-size function is as follows,
\[
f_{SCS}\left(\boldsymbol{x},G,\kappa\right)=\frac{\underset{i,j\in U_{\boldsymbol{x}},i\neq j}{\sum}\mathbb{I}\left[\left(i,j\right)\in E\right]}{\max\left(\left|U_{\boldsymbol{x}}\right|\left(\left|U_{\boldsymbol{x}}\right|-1+\kappa\right),1\right)}
\]

\subsection{\label{subsec:Optimization-Experimental-Results}Experimental Results}

As a benchmark for the clique problem, we used 80 undirected graphs
that were published as part of the second DIMACS challenge \cite{Johnson1996}.
Each graph was generated by a random generator that specializes in
a particular graph type that conceals cliques in a different manner.
The graphs contain up to 4000 nodes, and are varied both in their
number of nodes and in their edge density. We tested each method on
all 80 graphs, letting it maximize the soft-clique-size function using
various values of $\kappa$. Since a-priori we do not know which $\kappa$
will lead some method towards an inclusion maximal clique, we have
executed each method with each of the values $0.0,0.1,\dots,1.0$
as $\kappa$. We have executed a method for $100\left|V\right|$ samples
(hence runtime is fixed per graph), and recorded the following items
at the execution's end. We recorded the best solution that was found
during an execution, along with its objective value, and the sample
number in which that solution was found. 

In terms of the methods tested, following the discussion on related
work, we experimented with a version of CE, three versions of REINFORCE,
and of the bandit algorithms we used Exp3. As a sampling distribution,
we followed Rubinstein's construction that assumes independence between
the different dimensions, and we used $N$ softmax distributions defined
over the $N$ dimensions of a problem (the distribution is fully specified
in the supplementary material). Since we focus on online algorithms,
for CE, we derived a surrogate objective that causes the parameters
to update only when we encounter an example whose objective value
belongs to some $\text{\ensuremath{\rho}}$-highest percentile (also
specified in the supplementary material). For this surrogate, we get
an online algorithm that operates like CE with parameter smoothing
\cite{DeBoer2005}, and thus we refer to it as OCE with O standing
for online. We applied OCE with two values that were suggested by
Rubinstein, $\rho=0.1$ and $\rho=0.01$, and refer to these as $\text{OCE}_{0.1}$
and $\text{OCE}_{0.01}$. Next, we experimented with three versions
of REINFORCE. First is the vanilla version, second is a version where
the mean $\hat{\mu}$ is subtracted from $y$ as a baseline, and a
third uses the objective's estimated z-score $\frac{y-\hat{\mu}}{\hat{\sigma}}$.
We refer to these by $\text{REINF}$, $\text{REINF}_{B}$, and $\text{REINF}_{Z}$.
For Cakewalk, we used both the unscaled empirical CDF $\hat{F}$,
and its scaled counterpart $\hat{w}$, denoting these as $\text{CW}\,\hat{F}$
and $\text{CW}\,\hat{w}$ respectively. Note however that the former
is only used for comparison, and that we identify Cakewalk with the
latter. For estimating $\hat{\mu},\hat{\sigma}$ and $\hat{F}$, we
have used the last $100$ objective values. We emphasize that both
$\text{REINF}_{B}$, and $\text{REINF}_{Z}$ are important comparisons
as these methods only transform the objective values, but they do
not fix the distribution of the objective as CE and Cakewalk do. For
the gradient update steps, we have used vanilla stochastic gradient
ascent (SGA henceforth), AdaGrad, and the Adam gradient updates. The
latter two updates are considered scale invariant, and could therefore
help Exp3, $\text{REINF}$, and $\text{REINF}_{B}$ handle changes
in the objective's scale. Altogether, we have tested 8 adaptive samplers,
3 gradient updates, on 80 graphs, and 11 values of $\kappa$, leading
to a total of 21120 separate executions. We specify the complete experimental
details in the supplementary material. 

We analyzed 4 performance measures for each of the 8 samplers, and
the 3 gradient update types, and accordingly we report results in
four $3\times8$ tables. In the following, we refer to each combination
of a sampler and gradient update as a method. First, we examined whether
a locally optimal solution was found using a simple neighborhood function.
To that end, given a result $\boldsymbol{x}$ in some graph, we compared
$\boldsymbol{x}$ to every other $\boldsymbol{x}^{\prime}$ such that
$\sum_{i}\left|x_{i}-x_{i}^{\prime}\right|=1$, and checked that no
$\boldsymbol{x}'$ in that graph achieved higher soft-clique-size
than $\boldsymbol{x}$. We report the rates at which locally optimal
solutions were found in table \ref{tbl:Local-Optimality-SCS}. Then,
we proceeded to test if the returned solutions were inclusion maximal.
Since the soft-clique-size does not guarantee convergence to cliques,
for every graph, we tested whether a method returned at least one
inclusion maximal clique when applied with some $\kappa$. We report
the rates at which inclusion maximal cliques were found in table \ref{tbl:Local-Optimality-IncMax}.
Since some methods find their best solution earlier than others, to
analyze the sampling efficiency of each method we calculated the ratio
of the best-sample and the total-samples used in that execution. Since
this comparison only makes sense when controlling for the quality
of the solution, we excluded $\text{REINF}$ and Exp3 from it as they
did not return locally optimal solutions. We report average best-sample
to total-samples ratios in table \ref{tbl:Best-Sample-Ratio}. To
ensure returned solutions are not trivial (say cliques of size 2),
for each graph, we compared the largest inclusion maximal clique found
by that method, and compared it to the best known solution for that
graph, using results from \cite{DIMACS-MaxClique-Results}. We report
average largest-found-clique to largest-known-clique ratios in table
\ref{tbl:Larget-Clique-Ratio}. Lastly, we performed multiple hypothesis
tests to compare every sampler to $\text{CW}\,\hat{w}$ in all the
experimental conditions using one sided sign test \cite{Gibbons2011}.
To control the false discovery rate \cite{Wasserman2013}, we determined
the significance threshold at a level of $10^{-2}$ using the Benjamini-Hochberg
method \cite{Wasserman2013}. In all the tables in this section, when
a method is out performed in a statistically significant manner by
$\text{CW}\,\hat{w}$ this is denoted by $^{*}$. The best sampler
in each table is emphasized using bold fonts.

The results in tables \ref{tbl:Local-Optimality-SCS} and \ref{tbl:Local-Optimality-IncMax}
clearly support our main proposition: in our setting, a surrogate
objective function whose distribution is predetermined significantly
improves the robustness of a sampler, and accordingly improves the
rate at which locally optimal solutions are found. Both $\text{CW}\,\hat{w}$
and $\text{OCE}_{0.1}$ rely on such surrogates, and both outperform
Exp3 and all versions of REINFORCE which do not. Importantly, since
the previous comparison also includes $\text{REINF}_{Z}$, it follows
that it is not enough to simply normalize the objective values, and
it is better to actually fix the distribution of the objective. Nonetheless,
not all distributions are as effective ($\text{OCE}_{0.01}$ and $\text{CW}\,\hat{F}$
did not perform as well), and of the ones that we have tested, uniform
on $\left[-1,1\right]$ as used by $\text{CW}\,\hat{w}$ achieved
the best results. $\text{CW}\,\hat{w}$ clearly outperforms $\text{OCE}_{0.1}$
in table \ref{tbl:Local-Optimality-SCS}, and the latter only comes
close in the more permissive comparison which selects the best result
out of 11 different executions (different values of $\kappa$) as
reported in table \ref{tbl:Local-Optimality-IncMax}. In terms of
sampling efficiency, the results in table \ref{tbl:Best-Sample-Ratio}
show that even though $\text{OCE}_{0.1}$ can recover locally optimal
solutions, it is not as efficient as $\text{CW}\,\hat{w}$ which finds
the best solution considerably faster. When considering the various
gradient updates, $\text{CW}\,\hat{w}$ with AdaGrad produces the
best results in almost all measures ($\text{CW}\,\hat{w}$ with Adam
converges slightly faster, though at the cost of worse optimality
rates). This is unsurprising as AdaGrad's classical use case is sparse
data (indicator vectors in this case). Lastly, the comparisons to
the best known results in table \ref{tbl:Larget-Clique-Ratio} show
that the recovered solutions are far from trivial, and that Cakewalk
might even approach the performance of problem specific algorithms
which have access to a complete specification of the problem. 

\section{\label{sec:Application-to-K-Medoids}K-Medoid Clustering}

\begin{table*}[t]
	\caption{K-medoid clustering results}
	\label{tbl:kmedoids}
	\centering	
	\begin{tabular}{l|lllllll}
		\toprule
				   & Rank1  & P-Value   & Rank2     & P-Value   & Rank3     & P-Value   & Rank4     \\ 
		\midrule
		Objective  & CWV    & 0.003816$^{*}$  & PAM       & 7.276e-12$^{*}$ & CW        & 1.419e-10$^{*}$ & VOR       \\
		& 1.0002 &           & 1.0025    &           & 1.0426    &           & 1.6373    \\
		\midrule		
		Evaluations & VOR    & 3.638e-12$^{*}$ & CWV       & 3.638e-12$^{*}$ & CW        & 0.1279    & PAM       \\
		& 1.0000 &           & 1764.1380 &           & 4352.3860 &           & 4922.3370 \\ 
		\bottomrule
	\end{tabular}
\end{table*}

In this section we demonstrate how Cakewalk
can be used as a method that learns how to initialize greedy algorithms.
The key idea we utilize in this section is the following. Since Cakewalk
only relies on function evaluations, it does not matter if we let
it optimize some function $f:\mathcal{X}\to\mathbb{R}$, or a composition
$f\circ T$ where $T:\mathcal{X}^{\prime}\to\mathcal{X}$ is some
deterministic transformation ($\mathcal{X}^{\prime}$ can be identical
to $\mathcal{X}$, or some other space that specifies possible initializations
of a procedure $T$). As long as some input $\boldsymbol{x}$ is associated
with some fixed objective value $y=f\left(T\left(\boldsymbol{x}\right)\right)$,
Cakewalk will be able to optimize it. Thus, we can treat a deterministic
greedy algorithm as such $T,$ and use Cakewalk to optimize its initialization. 

To demonstrate such a usage we study the k-medoids \cite{Hastie2009}
problem, the combinatorial counterpart of k-means. As in the k-means,
we are given a set of  $m$ data points, and our goal is to divide
these into $k$ clusters which minimize the points' distances to a
set of cluster representatives. In contrast to k-means, in k-medoids
the representatives must be a subset of the original points that we
are given. Thus, one can think of the problem as selecting $k$ representatives
from the $m$ data points, and in the general case where we allow
points to represent more than one cluster, the solution space becomes
$\left[m\right]^{k}$. Since in k-medoids the representatives are
a subset of the data points, it is enough to consider as input a distance
matrix $D\in\mathbb{R}_{+}^{m\times m}$ where $D_{i,j}$ is the distance
between point $i$ and $j$, and $\mathbb{R}_{+}$ is the set of non-negative
reals. Given a set of representatives $\boldsymbol{x}\in\left[m\right]^{k}$,
each point $i$ is assigned to the representative $x_{j}$ which minimizes
the distance $D_{i,x_{j}}$ to it. In this formulation, the k-medoids
optimization problem can be stated as follows, 
\[
\begin{aligned} & \underset{x\in\left[m\right]^{k}}{\text{minimize}} &  & \sum_{i=1}^{m}\left[\underset{j\in\left[k\right]}{\min}\left\{ D_{i,x_{j}}\right\} \right]\end{aligned}
\]
\\
Since the problem is combinatorial, going over all the possible solutions
quickly becomes intractable, and greedy algorithms are usually used
to approach the problem. Of these, probably the two most commonly
used algorithms are the Voronoi iteration \cite{Hastie2009}, and
the more computationally expensive, Partitioning Around Medoids \cite{Kaufman2009}
(PAM henceforth).

\subsection{\label{subsec:K-Medoids-Experimental-Results}Experimental Results}

Using a publicly available collection \cite{RDatasets}, we gathered
38 datasets that had between 500 and 1000 data points. In each dataset
we extracted all the numerical attributes, and used these to represent
each data point. Then, for each dataset we calculated pairwise Mahalanobis
distances, using diagonal covariance matrices \cite{Bishop2006}.
At this point, we were able to test the aforementioned algorithms
on these datasets. Specifically, we used the Voronoi iteration and
PAM, as well as vanilla Cakewalk. As an example for a use case where
Cakewalk is combined with a greedy method, we also used Cakewalk with
the Voronoi iteration. We did not combine Cakewalk with PAM as it
is considerably more computationally expensive than the former. In
the result table which we specify next, we refer to these by VOR,
PAM, CW, and CWV. We applied Cakewalk using the best configuration
found in section \ref{subsec:Optimization-Experimental-Results},
and the same factorized distribution. We applied each of the 4 algorithms
on all datasets with $k=10$, and recorded the best objective value
that was returned, as well as the number of objective evaluations
that were performed. We specify the complete experimental details
in the supplementary material. 

In the analysis our goal was to produce two statistically significant
rankings of the methods tested. First, we ranked the methods by the
objective values they returned. Second, as the former criterion is
influenced by the number of objective function evaluations, we also
recorded the number of evaluations each method performed. To produce
rankings for both criteria, we compared the 4 methods in a manner
that is invariant to the specifics of a given dataset. For that matter,
we first calculated the ratio between a method's score (objective
value, or number of evaluations) in some dataset, and the minimal
score achieved by any method on that dataset. Then, we averaged these
ratios on all 38 datasets, and used that averaged ratio as a method's
score. This provided us with two scores for each of the 4 methods.
Sorting these averaged ratios provided us with a ranking which we
could then test for statistical significance. For each two consecutive
methods in a ranking, we performed one sided sign test using the original
values measured on the 38 datasets. This procedure produced 3 p-values
for the ranking of the objective values, and another 3 for the number
of evaluations. Lastly, to control the false discovery rate, we determined
the significance threshold at a level of $10^{-2}$ using the Benjamini-Hochberg
method. We report the results of this analysis in table \ref{tbl:kmedoids},
where rank1 one is the best method, and rank4 is the worst. In table
\ref{tbl:kmedoids}, the ranking in terms of objective values is displayed
in the first row, and the ranking in terms of number of evaluations
is displayed in the second row. For each ranking, presented are the
methods names, along with the p-values for the difference between
each pair. When the difference between a pair of methods is statistically
significant, this is denoted by $^{*}$. In addition, under each method
are the averaged ratios used to produce the ranking. 

These results demonstrate that combining Cakewalk with a greedy algorithm
can produce a method that outperforms the components that make it
up. Notably, here we combined Cakewalk with the Voronoi iteration,
the weaker of the two greedy methods we tested, and that already produced
the best results. Furthermore, vanilla Cakewalk outperformed the Voronoi
iteration, showing that Cakewalk can outperform some greedy algorithms
as these might be limited by the transformations they apply, and the
initializations they rely on. In terms of function evaluations, it
appears that providing Cakewalk with the Voronoi iteration leads to
faster convergence compared to vanilla Cakewalk, another positive
outcome for this combined optimization strategy. 

\section{\label{sec:Conclusion}Conclusion}

In this paper we presented Cakewalk, a generic adaptive sampler for
combinatorial optimization problems. We demonstrated how Cakewalk
outperforms similar adaptive samplers, and how Cakewalk can be combined
with greedy algorithms to produce highly effective optimizers. We
believe that future research will prove Cakewalk\textquoteright s
effectiveness in combinatorial problems that arise in practice, as
well as in other domains such as continuous non-convex optimization,
and reinforcement learning.

\section{Acknowledgment}

Supported by Israeli Science Foundation grant 320/16.

\bibliography{cakewalk}
\bibliographystyle{aaai}

\end{document}



\section{Sampling Distribution \label{subsec:Sampling-Distribution}}

Before we specify a particular distribution, we wish to emphasize
that Cakewalk is not specific to any particular sampling distribution.
One can construct sampling distributions that convey some prior knowledge
about a problem, and through such specification aid the optimization
process. Having said that, as a basis for the experiments we report
in the paper we utilize the following simple distribution. Following
Rubinstein's construction, we use a simple distribution that factorizes
into a sequence of independent distributions, each defined over a
different dimension. In this manner, the number of parameters required
to represent $\mathbb{P}_{\boldsymbol{\theta}}\left(\boldsymbol{x}\right)$
grows only linearly with $MN$, instead of the exponential number
of parameters that is required to represent the full joint distribution.
Formally, each $x_{j}$ is drawn independently according to a softmax
distribution $\mathbb{P}_{\boldsymbol{\theta}}\left(x_{j}=i\right)=\frac{e^{\theta_{i.j}}}{\sum_{k\in\left[M\right]}e^{\theta_{k,j}}}$
where $i\in\left[M\right]$, and therefore $\mathbb{P}_{\boldsymbol{\theta}}\left(\boldsymbol{x}\right)=\underset{j\in\left[N\right]}{\prod}\frac{e^{\theta_{x_{j},j}}}{\sum_{k\in\left[M\right]}e^{\theta_{k,j}}}$.
Next, we describe $\nabla_{\boldsymbol{\theta}}\log\mathbb{P}_{\boldsymbol{\theta}}\left(\boldsymbol{x}\right)$
in terms of partial derivatives, $\frac{\partial\log\mathbb{P}_{\boldsymbol{\theta}}\left(\boldsymbol{x}\right)}{\partial\theta_{i,j}}=\mathbb{I}\left[x_{j}=i\right]-\mathbb{P}_{\boldsymbol{\theta}}\left(x_{j}=i\right)$.
Note that since $\left\Vert \nabla_{\boldsymbol{\theta}}\log\mathbb{P}_{\boldsymbol{\theta}}\left(\boldsymbol{x}\right)\right\Vert $
is bounded, we can estimate $\hat{F}$ in an online manner. 

\section{Online Cross Entropy}

We present a surrogate objective that produces an online algorithm
that performs parameter updates that are similar to those of the CE
method. Recall that our surrogate objectives are weight functions
that are composed over the original objective $f$, where each weight
function is constructed using the empirical CDF $\hat{F}$ and another
monotonic transformation $g$. Thus, we denote this weight function
by $\hat{w}_{CE}$, and its associated transformation by $g_{CE}$.
Given some small $\rho\in\left[0,1\right]$ which is decided by the
user a-priori (typically, $0.1$ or $0.01$), $g_{CE}$ is a thresholding
function $g_{CE}\left(z\right)=\mathbb{I}\left[z\ge1-\rho\right]$,
and thus $\hat{w}_{CE}\left(y\right)=\mathbb{I}\left[\hat{F}\left(y\right)\ge1-\rho\right]$.
Clearly, for any fixed $\rho$, $\hat{w}_{CE}$ is monotonic increasing
and bounded, and $\hat{w}_{CE}\left(\boldsymbol{Y}\right)$ is a Bernoulli
random variable with probability $\rho$. Notably, using $\hat{w}_{CE}$
we get an algorithm that at each iteration $t$ samples an example
$\boldsymbol{x}_{t}$, receives an objective value $y_{t}=f\left(\boldsymbol{x}_{t}\right)$,
and updates the sampling distribution's parameters' $\boldsymbol{\theta}_{t}$
as follows,
\begin{align*}
\boldsymbol{\theta}_{t} & =\boldsymbol{\theta}_{t-1}+\eta\mathbb{I}\left[\hat{F}\left(y_{t}\right)\ge1-\rho\right]\nabla_{\boldsymbol{\theta}}\log\mathbb{P}_{\boldsymbol{\theta}}\left(\boldsymbol{x}_{t}\right)
\end{align*}
\\
Thus, if $y_{t}$ belongs to the highest $\rho$-percentile, we get
that $\mathbb{I}\left[\hat{F}\left(y_{t}\right)\ge1-\rho\right]=1$
and the parameters are updated. Since $\eta$ is positive, the parameters
are updated in a direction that increases $\boldsymbol{x_{t}}$'s
likelihood. Thus, we get an algorithm that increases the likelihood
of only the examples that belong to the highest $\rho$-percentile,
which is similar to the CE's update rule \cite{DeBoer2005}. Importantly,
in this formulation we get an online algorithm, and therefore we refer
to this algorithm as Online Cross Entropy. We emphasize however that
the convergence guarantees for OCE follow the policy gradient theorem
\cite{Sutton2017} rather than Rubinstein's work \cite{Rubinstein2001}. 

\section{\label{sec:Clique-Details}Maximum Clique Experiment Details }

As a benchmark, we used 80 undirected graphs that were published as
part of the second DIMACS challenge \cite{Johnson1996} which specifically
focused on combinatorial optimization, and included instances of the
clique problem. Over the years, this dataset has become a standard
benchmark for clique finding algorithms, and results on it are regularly
published. 

In terms of the methods we use for comparison, of the bandits family
of algorithms, we considered Exp3 as more suitable than UCB due to
the multi-dimensionality of the problem. For example, adding an isolated
vertex $v$ to a set of vertices $U$ who is a clique will damage
the objective. Due to such cases, we used Exp3 instead of UCB. We
applied Exp3 to each of the $N$ elements independently. Note that
the assumption of bounded losses/gains that the Exp3 algorithm is
dependent upon is met by the soft-clique-size function. 

For the gradient updates, we used SGA, AdaGrad, and Adam. We note
that AdaGrad is particularly suited to our setting as applying it
on indicator data is one of its classical use cases ($\mathbb{I}\left[x_{j}=i\right]$
can be considered as our data). Adam on the other hand has proven
as effective for training neural networks in a wide variety of problems,
and nowadays is probably the mostly commonly used gradient update.
We decided to experiment with Exp3 in conjunction with AdaGrad and
Adam even though this revokes the theoretical guarantees of Exp3 for
completeness purposes. We applied AdaGrad with $\delta=10^{-6}$,
and Adam with $\beta_{1}=0.9,\beta_{2}=0.999,\epsilon=10^{-6}$. We
used a fixed learning rate of $0.01$ in all the executions. All the
algorithms were implemented in Julia \cite{Bezanson2017} by the authors. 

\section{\label{sec:K-Medoids-Details}K-Medoids Experiment Details }

In the paper we compared Cakewalk to the Voronoi iteration \cite{Hastie2009},
and PAM \cite{Kaufman2009}, and showed how Cakewalk can be combined
with such greedy algorithms. Here we provide a short description of
the Voronoi iteration and PAM procedure. In both methods, first some
initial set of representatives is determined, and the appropriate
cluster assignments are determined. In the former method, in each
iteration, we seek to replace each representative with some other
cluster member, and in the latter we seek to replace each representative
with any non-representative point. After the new representatives are
determined, cluster assignments are determined, and the process is
then repeated as long as the objective is improved. In spite of the
obvious computational benefits of the Voronoi iteration, PAM is probably
more commonly used as it is known to achieve lower objective values. 

In terms of Cakewalk's configuration, we applied it with AdaGrad using
the same hyper-parameters as specified in section \ref{sec:Clique-Details},
except for the learning rate $\eta$, which was set to $0.02$ instead
of $0.01$ as we have seen that it led to faster convergence. We used
this configuration both when applying Cakewalk alone, and when applying
the Cakewalk-Voronoi combination. As a convergence criterion for Cakewalk,
we use two exponentially running averages of the objective values,
and determined convergence has occurred when their absolute difference
ratio was smaller than $0.01$. Each running average was produced
using a time constant that was calculated using the following formula:
$1-\exp\left(\frac{\ln\left(a\right)}{b}\right)$ with $a$ always
being $0.01$, and $b$ being a parameter that is adjusted to the
size of the problem. For each dataset with $m$ data points and $k$
clusters, we have set $b=\max\left(mk,1000\right)$ for the average
that corresponds to the smaller time window, and $2b$ for the average
that corresponds to the larger time window. We used the same converge
criterion for Cakewalk+Voronoi.

Next, we note that when considering the ranking in terms of the number
of function evaluations, one has to take into account the different
nature of Cakewalk+Voronoi. In the latter case, each evaluation corresponds
to an application of the Voronoi iteration, and thus, its number of
function  evaluations cannot be compared directly with the Voronoi
iteration, or with PAM. However, comparing it to Cakewalk does makes
for a valid comparison, and notably, the differences are statistically
significant showing that providing Cakewalk with a greedy algorithm
can speedup Cakewalk's convergence. 

\bibliography{cakewalk}
\bibliographystyle{aaai}